\documentclass[runningheads]{llncs}
\usepackage[T1]{fontenc}
\usepackage{graphicx}
\usepackage{booktabs}
\usepackage[misc]{ifsym}
\usepackage{algorithm}
\usepackage{algpseudocode}
\usepackage{amsmath}
\usepackage{amssymb}
\usepackage{subcaption}
\usepackage{longtable}
\usepackage{pdflscape}
\usepackage{pgfplots}
\usepackage{bm}
\usepackage{tikz}
\pgfplotsset{compat=newest}

\newcommand{\corr}{(\Letter)}
\usepackage{mwe}

\begin{document}

\title{Decentralized Time Series Classification with ROCKET Features}

\titlerunning{Decentralized Time Series Classification with ROCKET Features}

\author{Bruno Casella\inst{1}\orcidID{0000-0002-9513-6087} \corr \and
Matthias Jakobs\inst{2}\orcidID{0000-0003-4607-8957}  \and
Marco Aldinucci\inst{1}\orcidID{0000-0001-8788-0829} \and Sebastian Buschj\"ager\inst{2}\orcidID{0000-0002-2780-3618}}

\authorrunning{B. Casella et al.}

\institute{University of Turin, Turin, 10149, Italy \email{\{bruno.casella,marco.aldinucci\}@unito.it}
\and
Lamarr Institute for Machine Learning and Artificial Intelligence
TU Dortmund University, Dortmund, Germany \email{\{matthias.jakobs,sebastian.buschjaeger\}@tu-dortmund.de}
}

\maketitle              

\begin{abstract}
Time series classification (TSC) is a critical task with applications in various domains, including healthcare, finance, and industrial monitoring. Due to privacy concerns and data regulations, Federated Learning has emerged as a promising approach for learning from distributed time series data without centralizing raw information. However, most FL solutions rely on a client-server architecture, which introduces robustness and confidentiality risks related to the distinguished role of the server, which is a single point of failure and can observe knowledge extracted from clients. To address these challenges, we propose DROCKS, a fully decentralized FL framework for TSC that leverages ROCKET (RandOm Convolutional KErnel Transform) features. In DROCKS, the global model is trained by sequentially traversing a structured path across federation nodes, where each node refines the model and selects the most effective local kernels before passing them to the successor. Extensive experiments on the UCR archive demonstrate that DROCKS outperforms state-of-the-art client-server FL approaches while being more resilient to node failures and malicious attacks. Our code is available at \url{https://anonymous.4open.science/r/DROCKS-7FF3/README.md}.

\keywords{federated learning \and time series classification \and rocket \and decentralized learning.}
\end{abstract}

\section{Introduction}
\label{sec:introduction}
Time series classification (TSC) is popular in various domains due to the abundance of time series (TS) data in everyday activities. TSC has pivotal applications in various real-world scenarios, including healthcare~\cite{rajkomar2018scalable} (e.g., sleep stage classification from physiological data, or electrocardiogram classification), human activity recognition~\cite{nweke2018deep} and cyber-security~\cite{susto2018ts}. 

Given the increasing availability of TS data, developing efficient and accurate classification methods is crucial. While Deep Learning (DL) techniques have shown remarkable success in TSC, they come with high computational costs and energy consumption, making them impractical for many real-world applications.
In the case of TSC, it has been shown that quite simple algorithms, such as ROCKET~\cite{dempster2020rocket}, can achieve comparable performance while requiring significantly lower computational costs, leading to better energy efficiency~\cite{bagnall2017ts,pasosruiz2021ts}.
ROCKET (RandOm Convolutional KErnel Transform) utilizes a set of randomly sampled convolutional kernels to transform data, which is subsequently processed by a 
linear model to select significant features. 
This method not only achieves state-of-the-art accuracy while reducing the model size but, due to the random sampling, is also well-suited for low-resource environments since it omits the cost-intensive training of convolutional kernels. This makes it an efficient alternative to DL-based methods, particularly in distributed and edge computing settings.

Additionally, logically or physically centralizing distributed sensitive data for training AI models introduces privacy issues, such as the risk of unauthorized access, potential disclosure or breach of personal information, and loss of control over personal information during storage or transfer. 
Federated Learning (FL)~\cite{mcmaham2017communication} has emerged as an effective way to address these privacy issues by enabling collaborative training of AI models while keeping data local. 
In its original description ~\cite{mcmaham2017communication}, multiple parties (clients) collaborate in solving a learning task using their private data. 
Importantly, each client's data is not exchanged or transferred to any participant. 
Clients collaborate by exchanging local models via a central server (aggregator), which collects and aggregates the local models to produce a global model.
However, typical FL aggregation mechanisms employ gradient or parameter averaging, which can interfere with convergence by randomly merging useful and less important weights~\cite{zhao2018federated}.
Additionally, in common FL settings, the distinct role of the central server introduces side effects because it sets it up as a single point of failure in the system. This logical schema is often implemented using a master-worker paradigm; the master plays the role of the server/aggregator, whereas the clients behave as workers. 

The system's robustness is not the only issue; security concerns arise when the master is \textit{semi-honest}~\cite{evans2018pragmatic}, i.e., it might attempt to reconstruct original data from gradients.

To address these challenges, such as privacy risks, security vulnerabilities, and the drawbacks of aggregating heterogeneous model updates, recent works have explored alternative FL strategies beyond traditional DL and model aggregation. 
Specifically, 
FROCKS (Federated RandOm Convolutional KErnel Transform), for example, combines FL with ROCKET~\cite{dempster2020rocket}, allowing clients to share learned kernel features together with model parameters. 
However, FROCKS is still constrained by its reliance on a central server and is limited to binary classification tasks. 
Given these limitations, developing methods that address these challenges is crucial, especially considering the growing significance of FL and TSC.

In this work, we propose DROCKS, a fully decentralized FL approach for TSC that extends FROCKS by eliminating the need for a server and by supporting multiclass classification.
DROCKS employs a ring communication schema, where each node sequentially trains a local linear model and transmits it, along with the most effective ROCKET kernels, to the next node. 
The subsequent node then fine-tunes the model using both received kernels and newly generated random kernels. 
This decentralized method addresses the concerns arising from server-based methods, extends FROCKS's approach to multiclass classification problems, and improves performance on real-world TSC tasks.

We validate DROCKS through extensive experiments on 128 binary and multiclass classification datasets from the UCR archive. In our extensive experimental evaluation, we cover a wide range of configurations, testing different numbers of ROCKET kernels to assess their impact on performance, exploring various decentralized topologies to analyze the robustness of the method, and investigating scalability across an increasing number of clients. The results demonstrate the superiority of DROCKS over state-of-the-art methods in terms of F1 score, with minimal computational and communication overheads.

\section{Related work}
\label{sec:related}
In this section, we start by presenting the state-of-the-art work on TSC before discussing why FL is important for TSC and how existing TSC algorithms can be extended to the federated setting.

Some of the most accurate TS classifiers are dictionary-based and ensemble learning methods. 
Bag-of-SFA-Symbols (BOSS)~\cite{schafer2015boss} achieves strong classification performance through Symbolic Fourier approximation but has a quadratic training complexity. 
Scalable variants like BOSS-VS reduce complexity at the cost of accuracy.
Shapelet-based methods~\cite{bagnall2017ts} extract discriminative subseries, providing high accuracy at the cost of quartic complexity in TS length.
Hierarchical Vote Collective of Transformation-Based Ensembles (HIVE-COTE)~\cite{lines2016hivecote}, an ensemble including BOSS and shapelets, achieves high accuracy but remains computationally expensive.

DL approaches have emerged as powerful alternatives, leveraging the sequential structure of TS data.
U-Time~\cite{perslev2019utime} is a temporal convolutional neural network (CNN) based on the U-Net~\cite{ronneberger2015unet} architecture, originally proposed for image segmentation. It classifies each time point and aggregates predictions over intervals. 
InceptionTime~\cite{fawaz2020inceptiontime}, an ensemble of Inception~\cite{szegedy2015inception} modules, achieves competitive results and benefits from efficient training via SGD with linear complexity. 

More recently, ROCKET~\cite{dempster2020rocket} demonstrated that extracting features using random convolutional kernels enables fast training while maintaining state-of-the-art classification performance.

The emergence of these methods has significantly contributed to advancements in TSC. However, as TSC plays a crucial role in domains like healthcare, finance, and cybersecurity, sensitive and private data are often involved.  
This sensitivity demands privacy-preserving AI approaches that maintain high classification performance.

FL, introduced with the FederatedAveraging (FedAvg)~\cite{mcmaham2017communication} algorithm, enables distributed learning by aggregating local updates while keeping data decentralized.

FL for TSC has seen strong recent advancements. FedTSC~\cite{liang2022fedtsc} is a federated TSC solution focusing on model interpretability based on explainable features.
FedST~\cite{liang2023fedst} extends FedTSC by elaborating on the design ideas and essential techniques of a main internal of the system. FedST introduces a secure federated shapelet transformation method. Shapelets, representing discriminative subsequences of TS data to identify classes, are extracted in a privacy-preserving manner, thus providing an efficient discovery across distributed datasets while ensuring security.

More recently, FROCKS~\cite{casella24frocks} proposes a federated TSC method based on ROCKET features. ROCKET feeds a linear classifier with data transformed with random convolutional kernels, thus providing superior speed without any drop in classification performance. FROCKS adapts ROCKET to a federated setting by distributing and selecting the best-performing set of kernels.
However, FROCKS is limited to only binary classification tasks and suffers from the single point of failure problem due to the client-server nature. 

Several decentralized alternatives have been explored to overcome the single point of failure limitation typical of standard FL architectures.
Blockchain-based approaches, such as BAFFLE~\cite{ramanan2020baffle} and VBFL~\cite{chen2021robust}, decentralize the aggregation process and enhance security by ensuring that model updates are validated and recorded in a tamper-proof ledger.
Another recent work, although primarily applied to image classification, proposes Fed\textit{ER}~\cite{pennisi2024feder}, a strategy exploiting experience replay and generative adversarial concepts with peer-to-peer communication between clients replacing the central server.

Another decentralized alternative to classical FL is gossip learning~\cite{gossip2019hegedus}, where nodes train local models independently and periodically exchange parameters with random peers. This process gradually converges toward a global model across the network, providing scalability and fault tolerance benefits. However, random peer-to-peer communication requires frequent exchanges between nodes, thus leading to significant network overhead. In contrast, our method requires less communication and computation, as a single model is iteratively and sequentially trained on each node for further fine-tuning.

In this work, we propose DROCKS, a decentralized FL method for TSC. We extend FROCKS to the multiclass classification task and address the single point of failure with a decentralized pipeline communication schema.

\section{Method}
\label{sec:method}

Before describing our proposed method, we briefly introduce the TSC setting and the necessary notation. 
We consider a setting of supervised univariate TSC, in which the goal is to learn a model that assigns a class label to an input TS based on observed patterns. 
Formally, each instance consists of a TS $\bm{x} = [ x_1, x_2, \dots, x_T ]$ of length $T$ and an associated class label $y \in \mathcal{C}$, where $\mathcal{C}$ is a finite set of possible classes. 
The challenge in TSC lies in effectively capturing temporal dependencies and discriminative features within the sequences. 
We will denote with $x_i$ the $i$-th value of time-series $\bm{x}$ and we will use $\bm{x}_{i:j}$ to denote the subseries $[x_i, x_{i+1}, \dots, x_{j}]$.
Next, let $\bm{x} * \bm{w}$ be the convolution of $\bm{x}$ with some other time-series $\bm{w} \in \mathbb{R}^L$, which we will refer to as the \textit{kernel}.
Convolution can be seen as computing the sliding dot-product between slices of $\bm{x}$ and the kernel $\bm{w}$ and is given by
$$
\bm{x} * \bm{w} = \left[\bm{w}^T\bm{x}_{1:L+1}, \bm{w}^T\bm{x}_{2:L+2}, \dots, \bm{w}^T\bm{x}_{(T-L):T} \right] 
$$

In convolutional neural networks, each layer consists of multiple kernels learned jointly with the rest of the network.
ROCKET\cite{dempster2020rocket}, on the other hand, is a feature extraction approach that generates a large number of kernels randomly and does not optimize them further.
In the original paper, the authors propose to use up to $K=10,000$ kernels for the best results.
After convolving the input TS with all generated kernels, ROCKET extracts two statistics per kernel: the maximum value of the resulting feature maps and the Percentage of Positive Values (PPV), measuring the number of times the dot products exceed zero. 
These extracted features are fed into a linear classifier.
In our work, we focus on the PPV features, since selecting both the maximum value of each kernel’s output and PPV has not been shown to provide a statistically significant advantage over using PPV alone~\cite{dempster2020rocket}. 

One of the first attempts at federating the ROCKET algorithm is FROCKS~\cite{casella24frocks}.
In FROCKS, each party of the federation starts training with a different set of ROCKET kernels. 
FROCKS trains a logistic regression on features obtained via ROCKET kernels. 
In particular, when dealing with a binary classification task, a logistic regression is made of a single vector of weights $\bm{\beta}$ of length $K$ (one weight per kernel, as FROCKS uses only PPV). 
If $\bm{x}$ is the original TS and we assume a set of kernels $\mathcal{W}$ with $|\mathcal{W}| = K$, we can denote the ROCKET transformation as $$\phi_{\mathcal{W}}(\bm{x}) = (\text{PPV}(\bm{x}*\bm{w}_1), \dots, \text{PPV}(\bm{x}*\bm{w}_K))^T.$$
For notational convenience, we define the transformation for a set of $M$ time-series $\mathcal{X} = \{ \bm{x}_i \}_{i=1}^M$ as $\phi_{\mathcal{W}}(\mathcal{X}) := \{ \phi_{\mathcal{W}}(\bm{x}_i) \}_{i=1}^M$.
After training a logistic regression, each client collects the $p$ best-performing kernels in terms of absolute weight $|\beta_i|^2$ (with $p = \left \lfloor \frac{K}{N} \right \rfloor$ where $N$ are the number of clients in the federation). 
The server gathers those kernels and their associated weights and builds a new set of kernels. If two clients send the same kernel, the server averages the corresponding weight. Before the next round of training, the new set of kernels is used to transform local data. FROCKS outperforms state-of-the-art methods and requires just a few rounds of training until convergence. 
Algorithm~\ref{alg:frocks}, in Appendix~\ref{secA4:frocks_multiclass}, describes the FROCKS algorithm.

However, FROCKS is inherently designed for binary classification and does not support multiclass tasks at all, as the algorithm (see Appendix~\ref{secA4:frocks_multiclass}) does not explicitly provide a mechanism for handling multiple classes. The natural and naive possible extension of FROCKS to the multiclass setting would be to train a separate binary classifier for each class and aggregate their outputs. However, this solution performs poorly on multiclass tasks. 
This is probably due to its intrinsic mechanism of merging/averaging kernels/weights that refer to different features. 
We hypothesize that, in a multiclass scenario, it is plausible that each classifier learns distinct features. As a result, averaging the weights associated with different features may lead to a degradation in learning performance. This occurs because, in multiclass tasks, the averaging process involves weights from classifiers that have learned potentially heterogeneous features, which can impact the overall learning outcome compared to binary classification, which involves a single classifier with a corresponding set of weights.
Moreover, it can exacerbate the problem of objective inconsistency~\cite{wang2020tackling}: standard averaging of client models after heterogeneous local updates may result in convergence to a stationary point, not of the original objective function $\mathcal{F}(x)$, 
but of an inconsistent objective $\tilde{\mathcal{F}}(x)$ which can be arbitrarily different from $\mathcal{F}(x)$ depending upon the relative values of local updates.

\begin{figure}[t]
\centering
\includegraphics[width=0.8\textwidth, trim=0cm 3cm 0cm 3cm]{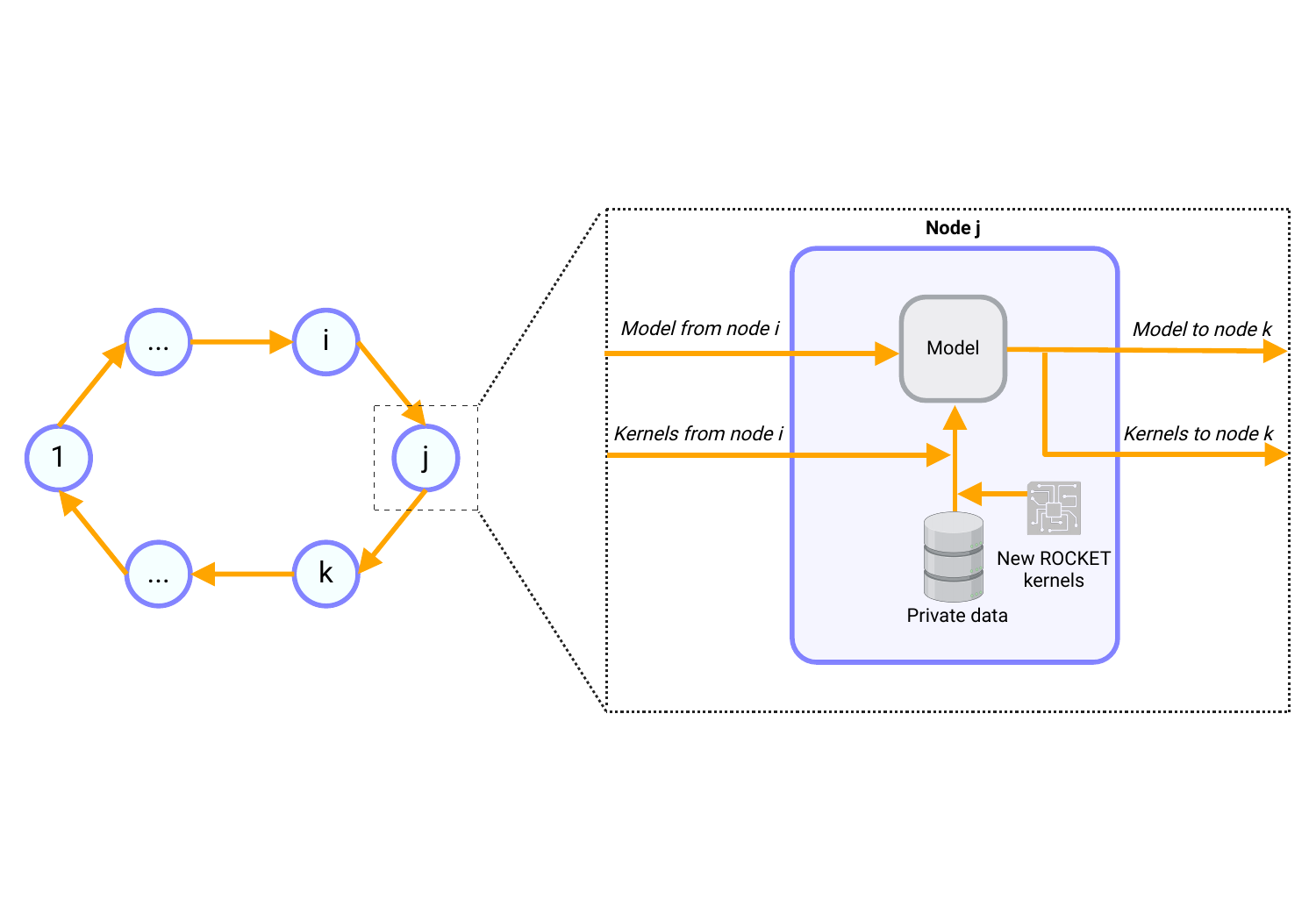} 
\caption{Each node in the sequence receives the trained model and the $p$ best-performing kernels from the preceding node. The node then fine-tunes the received model using its private data, transformed with a new set of ROCKET kernels that combine new kernels with the received ones.
}
\label{fig:overview}
\end{figure}

DROCKS extends FROCKS's principle of kernel selection to the multiclass scenario while removing the requirement of a centralized server for synchronization.
An overview of DROCKS is shown in Fig. \ref{fig:overview}. Contrary to FROCKS, we arrange the $N$ clients of the federation in a ring communication scheme where we communicate weights and kernels to the next client in line. This cyclical weight transfer has shown equal performance to centralized training while mitigating the need for a central server ~\cite{chang2018distributed}.  Training proceeds as follows:
Before training starts, each federation party samples its own set of $K$ Rocket kernels. 
In the first round of training, one of the clients is selected as the first node. It initializes the parameters of a linear model and trains it on its local data transformed with $K$ kernels. 
After the training phase, the first client selects the $p$ best-performing ROCKET kernels, with $p= \left \lfloor \frac{K}{N} \right \rfloor $, determined according to the largest squared weights associated with each kernel, and sends the trained model along with those selected kernels to the subsequent node of the federation. 
Upon receiving the model and the selected kernels, the new node transforms its TS data using a set of kernels comprising the received ones and $K-p$ new random kernels and retrains the model. 
Finally, the node selects a new set of $p$ best-performing kernels and forwards the updated model and kernel set to the next client in the sequence. 
In this way, each client will always operate with $K$ kernels.
Since no parameter averaging is involved in the process, this approach limits the performance degradation due to the aggregation of heterogeneous features in multiclass scenarios.
The process is iterative and continues through all the $N$ clients in the network, forming a chain of communication where each node refines the model with its distinct data and selected kernels. 
The cycle is repeated for several rounds of training. 
One round is considered completed when all clients sequentially train the model and select the best kernels.

The algorithm converges if the set of $p$ kernels with the largest weight is the same for two consecutive rounds.
Algorithm~\ref{alg:drocks} describes the DROCKS algorithm.

\begin{algorithm}
\caption{DROCKS: Decentralized ROCKET featureS}
\label{alg:drocks}
\begin{algorithmic}[1]
\Require $N$: number of clients, $\{D_i = (\mathcal{X}_i, \mathcal{Y}_i)\}_{i=1}^N$: local datasets, where $\mathcal{X}_i$ are inputs and $\mathcal{Y}_i$ are labels for client $i$, $K$: number of ROCKET kernels for each client, $R$: number of training rounds

\State Sample $p$ random kernels: $\mathcal{W} = \{\bm{w}_{j}\}_{j=1}^{p}$
\State Initialize linear model weights: $\bm{\beta} \in \mathbb{R}^K$

\For{round $r \in \{ 1, \dots R\}$}
    \For{client $i \in \{1, \dots, N\}$}

        \State Sample $K-p$ random kernels: $\mathcal{W}' = \{\bm{w}_{j}\}_{j=1}^{K-p}$
        \State Combine weights: $\mathcal{W}' = \mathcal{W} \cup \mathcal{W}'$ 
        \State Transform local data: $\mathcal{X}'_i = \phi_{\mathcal{W}'}(\mathcal{X}_i)$  
        \State Fine-tune logistic regression: $\bm{\beta}_i = \text{fit}(\mathcal{X}_i', \mathcal{Y}_i, \bm{\beta})$
        \State $\mathcal{W} = \{\mathcal{W}'_k: |\beta_k| ~ \text{is among top-k }  \}$
        \State $\bm{\beta} = \{\beta_k: |\beta_k| ~ \text{is among top-k }  \}$
        
        \State Send $\mathcal{W}, \bm{\beta}$ to the next client

    \EndFor
\EndFor

\State Stop training when $\mathcal{W},\bm{\beta}\footnotemark$ do not change from round to round

\end{algorithmic}
\end{algorithm}
\footnotetext{We use $\left|\bm{\beta}^{(r-1)} - \bm{\beta}^{(r)}\right| \le 10^{-8} + 10^{-5} \cdot \left|\bm{\beta}^{(r)}\right|$ to detect if there is a sufficient change between round $r$ and $r-1$.}

By iterating through the nodes and collaboratively selecting the most important kernels, DROCKS ensures that the model benefits from the diversity of the data distributed across the federation. Kernel selection is essential for learning distributed features and transferring them between nodes. The shared model extracts new features without forgetting the previously learned knowledge, thus incorporating continual learning principles and leading to a robust generalized model.

Unlike centralized FL methods, DROCKS does not rely on a central server to aggregate local models. This architectural choice eliminates the single point of failure represented by the server and reduces the risk of exposing the federation to privacy attacks by ``semi-honest''~\cite{evans2018pragmatic} clients attempting to reconstruct original data. However, the sequential nature of the pipeline topology introduces a new vulnerability, as each client could become a single failure point. If a client is compromised, disconnected, or fails during the training process, it can tamper with the model's propagation, impacting the overall cycle. To address these challenges and ensure the robustness of DROCKS, we identify two strategies that mitigate faults during training.
First, the ring topology can be replaced with a random topology, where clients receive and fine-tune the model from a random predecessor, enhancing fault tolerance.

In this configuration, a training round is considered complete only when all clients have participated, thereby enhancing fault tolerance and reducing dependence on a strictly sequential structure. 
Second, problematic clients, whether they are ``semi-honest'' adversaries or clients experiencing technical failures, can be excluded from the federation. 
By dynamically adapting the federation's composition, DROCKS ensures the continuity and security of the collaborative process.
These two strategies can be combined if a random topology is needed in the presence of compromised clients. The results of these settings are shown in Table~\ref{tab:abl:malicious_clients} and discussed in Section ~\ref{ssec:discussion}.


An additional advantage of this approach is its communication efficiency. 
First, the overall communication cost is limited by sending just $p$ kernels. Explicitly transmitting all kernels would be computationally expensive and significantly increase communication costs due to the large amount of kernels and their corresponding parameters. However, since the kernels are randomly generated, it suffices to exchange the random seed corresponding to each kernel. This seed can be used to recreate the exact parameters of the kernel when needed without actually sending the full parameter set. Thus, the DROCKS communication overhead depends only on the model's size and does not suffer from transmitting these kernels, as it involves only an integer rather than arrays of floating-point numbers. 
Second, unlike traditional FL server-based methods such as FedAvg, where each round involves dual communication (each client sends model parameters to the server, and the server returns the aggregated model), the proposed pipeline schema halves the communication overhead. Indeed, if $S$ represents the size of the model parameters, the overall communication cost per round in server-based approaches is $2 \cdot N \cdot S$, while DROCKS requires only $N \cdot S$, as it involves only a single communication per round, directly between clients.

\section{Experiments and results}
\label{sec:experiments}
In this section, we describe DROCKS's predictive capabilities compared to several baselines, the environmental setting, and the datasets and models used.

\subsection{Baselines}
We compare the performance of DROCKS with that of five different competitors.
\begin{itemize}
    \item \textbf{FedAvg-RawData} (abbreviated as RawData). In the simplest federated approach serving as a baseline, we train a logistic regression without using ROCKET features. Each client trains the logistic regression on its own data, and periodically, the server gathers the updates in a global model. The aggregation technique selected is the state-of-the-art algorithm FedAvg. Training lasts $R$ rounds.
    
    \item \textbf{FedAvg-ResNet-18} (abbreviated as ResNet). As in the \textit{RawData} baseline, here we train a time-series version of ResNet~\cite{he2016resnet} using 1D convolutions with the original data without transforming it with ROCKET kernels. We decided to train a ResNet as it is a baseline DL model for TSC and is widely adopted in FL settings. The FL algorithm is FedAvg, and training lasts $R$ rounds.
    
    \item \textbf{FedAvg-InceptionTime} (abbreviated as InceptionTime)~\cite{fawaz2020inceptiontime}. As for the \textit{RawData} and \textit{ResNet} baselines, we train an InceptionTime model on the raw data. We decided to train an InceptionTime as it is a state-of-the-art DL model for TSC, and its intrinsic gradient descent-based nature fits well with the federated process. FedAvg is adopted as an FL algorithm, and training lasts \textit{R} rounds.   
    
    \item \textbf{FedAvg-RocketFeatures} (abbreviated as RocketFL). In this setting, we train a logistic regression using ROCKET features. Before the federated training starts, the central server broadcasts the same set of ROCKET kernels to all the clients together with an initialized linear classifier. The clients will use the received kernels to extract features from their local data. After the transform phase, the typical federation training, as in the \textit{RawData}, begins. The aggregated model is trained for $R$ rounds with FedAvg.
    
    \item \textbf{FROCKS.} This competitor works well only on binary classification problems. 
    If kernels and weights do not change for two consecutive rounds, the overall approach is converged, and training is stopped.   
\end{itemize}

\subsection{Testbed setup}
The \textit{RawData}, \textit{RocketFL}, \textit{ResNet}, and \textit{InceptionTime} baseline experiments have been executed in a real distributed environment encompassing one server and four clients. Each entity is deployed on a dedicated server with an Intel\textregistered Xeon\textregistered processor (Skylake, IBRS, 8 sockets of one core) and one Tesla T4 GPU. To conduct our experiments, we adopted OpenFL~\cite{openfl2022foley}, an FL library that is Deep Learning framework-agnostic. We used PyTorch to train the models.

To ensure an unbiased evaluation, we utilize the official code repository\footnote{https://github.com/MatthiasJakobs/FROCKS} to reproduce FROCKS experiments. FROCKS ran a simulated federation with one server and $N=4$ clients on a machine with the previously listed hardware specification. FROCKS used Scikit-learn as the library for training the logistic regression.

DROCKS ran a simulated federation with one server and $N=4$ clients on a machine with the same hardware specification as FROCKS. It used Scikit-learn as the library to train the linear classifier. 
DROCKS can also be deployed in a real distributed environment using the StreamFlow framework~\cite{colonnelli2021streamflow}, a container-native workflow management system, to appreciate DROCKS' fault tolerance characteristics fully. This integration can enhance DROCKS's ability to handle faults, as StreamFlow's fault tolerance mechanisms~\cite{mulone2024fault} are well-suited for mitigating issues caused by compromised or failing clients. Specifically, StreamFlow can help maintain the continuity of the decentralized process by managing client failures dynamically, ensuring that the federation remains robust and operational despite potential disruptions. 
A StreamFlow implementation of DROCKS is available at~\url{https://anonymous.4open.science/r/DROCKS\_StreamFlow-E09C/README.md}.

\subsection{Datasets and models}
We tested DROCKS, FROCKS, RawData, RocketFL, ResNet, and the InceptionTime approaches on all the TSC datasets of the UCR archive~\cite{UCRArchive2018}. 
Specifically, the UCR archive encompasses 42 binary datasets and 86 multiclass datasets. The time series lengths across these datasets vary considerably, with the 'SmoothSubspace' dataset having the shortest series of length 15, representing the minimum in the archive, and the 'Rock' dataset featuring the longest series with a length of 2844, the maximum in the archive.
Also, the sizes of the training and test sets vary significantly across the archive. For example, the 'DiatomSizeReduction' dataset features a training set of only 16 samples—the smallest in the archive—while the 'ElectricDevices' dataset boasts a training set of 8926 samples, the largest in the archive.

The RawData, the RocketFL approach, and DROCKS use the same hyperparameters of the original ROCKET paper~\cite{dempster2020rocket}: we train a logistic regression minimizing the cross-entropy loss, using the Adam optimizer with a learning rate of $10^{-3}$. 
During preliminary experiments, we tested different batch sizes from ${\{2, 4, 8\}}$ and chose the one that showed the best performance.
The maximum number of training rounds $R$ was fixed to 100. However, both DROCKS and FROCKS required fewer iterations thanks to their convergence methods. 

We have run experiments with $K \in \{100, 500, 1\,000, 5\,000, 10\,000\}$ ROCKET kernels.
We split the dataset into training and testing data and distributed it to each client. The training set is used to fit the model, while the testing set is used for the inference phase. Since our method does not maintain $N$ local models (one per client) and an aggregated model, but instead, each node sequentially contributes to the training process, the final shared model is tested on the test set of each client.
The data is independently and identically distributed (\textit{i.i.d.}) across all clients. This means that each client's dataset is drawn from the same probability distribution, and each data point is statistically independent of the others.
We repeated each experiment five times using different random seeds, and we reported the average outcomes. 
In our comparison, we will concentrate on the F1 score (macro-averaged) because most of the datasets are imbalanced. 
Additional results, such as the top-1 accuracy, are available in Appendices~\ref{secA1} and~\ref{secA2}.

\subsection{Discussion}
\label{ssec:discussion}

Due to space constraints, in this section, we present only the critical difference diagrams, a powerful tool to compare outcomes of multiple treatments over multiple observations (a lower rank more to the right is better), and an excerpt of the f1-score values.
All the numerical results are reported in the Appendices~\ref{secA1} and~\ref{secA2}.

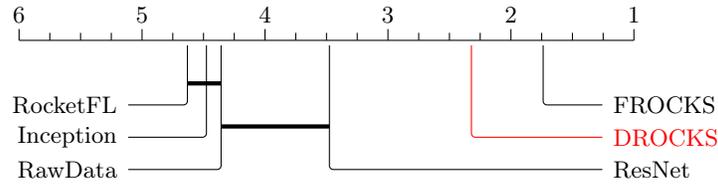
\begin{figure}
\centering
\begin{tikzpicture}[
  treatment line/.style={rounded corners=1.5pt, line cap=round, shorten >=1pt},
  treatment label/.style={font=\small},
  group line/.style={ultra thick},
]
\begin{axis}[
  clip={false},
  axis x line={center},
  axis y line={none},
  axis line style={-},
  xmin={1},
  ymax={0},
  scale only axis={true},
  width={0.67\columnwidth},
  ticklabel style={anchor=south, yshift=1.3*\pgfkeysvalueof{/pgfplots/major tick length}, font=\small},
  every tick/.style={draw=black},
  major tick style={yshift=.5*\pgfkeysvalueof{/pgfplots/major tick length}},
  minor tick style={yshift=.5*\pgfkeysvalueof{/pgfplots/minor tick length}},
  title style={yshift=\baselineskip},
  xmax={6},
  ymin={-4.5},
  height={5\baselineskip},
  xtick={1,2,3,4,5,6},
  minor x tick num={3},
  x dir={reverse},
]
\draw[treatment line] ([yshift=-2pt] axis cs:1.7380952380952381, 0) |- (axis cs:1.2380952380952381, -2.0)
  node[treatment label, anchor=west] {FROCKS};
\draw[color=red, treatment line] ([yshift=-2pt] axis cs:2.3214285714285716, 0) |- (axis cs:1.2380952380952381, -3.0)
  node[treatment label, anchor=west] {DROCKS};
\draw[treatment line] ([yshift=-2pt] axis cs:3.4761904761904763, 0) |- (axis cs:1.2380952380952381, -4.0)
  node[treatment label, anchor=west] {ResNet};
\draw[treatment line] ([yshift=-2pt] axis cs:4.357142857142857, 0) |- (axis cs:5.130952380952381, -4.0)
  node[treatment label, anchor=east] {RawData};
\draw[treatment line] ([yshift=-2pt] axis cs:4.476190476190476, 0) |- (axis cs:5.130952380952381, -3.0)
  node[treatment label, anchor=east] {Inception};
\draw[treatment line] ([yshift=-2pt] axis cs:4.630952380952381, 0) |- (axis cs:5.130952380952381, -2.0)
  node[treatment label, anchor=east] {RocketFL};
\draw[group line] (axis cs:4.357142857142857, -1.3333333333333333) -- (axis cs:4.630952380952381, -1.3333333333333333);
\draw[group line] (axis cs:3.4761904761904763, -2.6666666666666665) -- (axis cs:4.357142857142857, -2.6666666666666665);
\end{axis}
\end{tikzpicture}
\caption{Mean ranks for the competitors and DROCKS on binary classification tasks, each with the best hyperparameter.}
\label{fig:critdd_binary}
\end{figure}

\vspace{-30pt} 

\begin{figure}
    \centering
    \begin{tikzpicture}[
  treatment line/.style={rounded corners=1.5pt, line cap=round, shorten >=1pt},
  treatment label/.style={font=\small},
  group line/.style={ultra thick},
]

\begin{axis}[
  clip={false},
  axis x line={center},
  axis y line={none},
  axis line style={-},
  xmin={1},
  ymax={0},
  scale only axis={true},
  width={0.67\columnwidth},
  ticklabel style={anchor=south, yshift=1.3*\pgfkeysvalueof{/pgfplots/major tick length}, font=\small},
  every tick/.style={draw=black},
  major tick style={yshift=.5*\pgfkeysvalueof{/pgfplots/major tick length}},
  minor tick style={yshift=.5*\pgfkeysvalueof{/pgfplots/minor tick length}},
  title style={yshift=\baselineskip},
  xmax={6},
  ymin={-4.5},
  height={5\baselineskip},
  xtick={1,2,3,4,5,6},
  minor x tick num={3},
  x dir={reverse},
]

\draw[color=red, treatment line] ([yshift=-2pt] axis cs:2.197674418604651, 0) |- (axis cs:1.697674418604651, -2.0)
  node[treatment label, anchor=west] {DROCKS};
\draw[treatment line] ([yshift=-2pt] axis cs:2.953488372093023, 0) |- (axis cs:1.697674418604651, -3.0)
  node[treatment label, anchor=west] {ResNet};
\draw[treatment line] ([yshift=-2pt] axis cs:3.1686046511627906, 0) |- (axis cs:1.697674418604651, -4.0)
  node[treatment label, anchor=west] {RawData};
\draw[treatment line] ([yshift=-2pt] axis cs:3.2151162790697674, 0) |- (axis cs:6.453488372093023, -4.0)
  node[treatment label, anchor=east] {RocketFL};
\draw[treatment line] ([yshift=-2pt] axis cs:3.511627906976744, 0) |- (axis cs:6.453488372093023, -3.0)
  node[treatment label, anchor=east] {Inception};
\draw[treatment line] ([yshift=-2pt] axis cs:5.953488372093023, 0) |- (axis cs:6.453488372093023, -2.0)
  node[treatment label, anchor=east] {FROCKS};
\draw[group line] (axis cs:3.1686046511627906, -2.0) -- (axis cs:3.511627906976744, -2.0);
\draw[group line] (axis cs:2.953488372093023, -2.2) -- (axis cs:3.2151162790697674, -2.2);

\end{axis}
\end{tikzpicture}
    \caption{Mean ranks for the competitors and DROCKS on multiclass classification tasks, each with the best hyperparameters.}
    \label{fig:critdd_multiclass}
\end{figure}
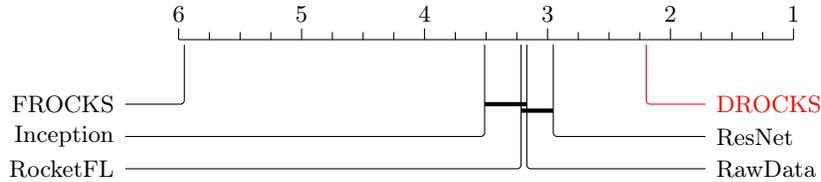

\vspace{-30pt} 

\begin{figure}
\centering
\begin{tikzpicture}[
  treatment line/.style={rounded corners=1.5pt, line cap=round, shorten >=1pt},
  treatment label/.style={font=\small},
  group line/.style={ultra thick},
]

\begin{axis}[
  clip={false},
  axis x line={center},
  axis y line={none},
  axis line style={-},
  xmin={1},
  ymax={0},
  scale only axis={true},
  width={0.67\columnwidth},
  ticklabel style={anchor=south, yshift=1.3*\pgfkeysvalueof{/pgfplots/major tick length}, font=\small},
  every tick/.style={draw=black},
  major tick style={yshift=.5*\pgfkeysvalueof{/pgfplots/major tick length}},
  minor tick style={yshift=.5*\pgfkeysvalueof{/pgfplots/minor tick length}},
  title style={yshift=\baselineskip},
  xmax={5},
  ymin={-3.5},
  height={4\baselineskip},
  xtick={1,2,3,4,5},
  minor x tick num={3},
  x dir={reverse},
]

\draw[treatment line] ([yshift=-2pt] axis cs:2.3046875, 0) |- (axis cs:1.8880208333333333, -2.5)
  node[treatment label, anchor=west] {K = 5000};
\draw[treatment line] ([yshift=-2pt] axis cs:2.6328125, 0) |- (axis cs:1.8880208333333333, -3.5)
  node[treatment label, anchor=west] {K = 10000};
\draw[treatment line] ([yshift=-2pt] axis cs:2.765625, 0) |- (axis cs:4.514322916666667, -4.0)
  node[treatment label, anchor=east] {K = 1000};
\draw[treatment line] ([yshift=-2pt] axis cs:3.19921875, 0) |- (axis cs:4.514322916666667, -3.0)
  node[treatment label, anchor=east] {K = 500};
\draw[treatment line] ([yshift=-2pt] axis cs:4.09765625, 0) |- (axis cs:4.514322916666667, -2.0)
  node[treatment label, anchor=east] {K = 100};
\draw[group line] (axis cs:2.6328125, -2.3333333333333335) -- (axis cs:2.765625, -2.3333333333333335);
\draw[group line] (axis cs:2.3046875, -1.6666666666666667) -- (axis cs:2.6328125, -1.6666666666666667);

\end{axis}
\end{tikzpicture}
\caption{Mean ranks of DROCKS with different kernel counts across all UCR datasets.}
    \label{fig:critdd_drocks}
\end{figure}
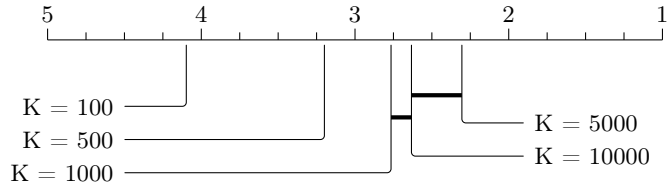

\begin{table}[!ht]
\centering
\caption{Excerpt of the numerical results (f1-scores).}
\label{tab:excerpt}
\setlength{\tabcolsep}{0.6mm} 
\resizebox{.99\columnwidth}{!}{
\begin{tabular}{lccccccc}
\hline
Dataset & \# classes & RawData & ResNet & InceptioTime & RocketFL & FROCKS & DROCKS \\
 \hline   
        Adiac & 37 & 0.509 & 0.459 & 0.319 & 0.376 & 0.000 & \textbf{0.652} \\
        Chinatown & 2 & 0.725 & 0.726 & 0.724 & 0.724 & \textbf{0.982} & 0.923 \\
        Crop & 24 & 0.399 & 0.516 & 0.428 & 0.367 & 0.001 & \textbf{0.595} \\
        ItalyPowerDemand & 2 & 0.908 & 0.929 & 0.883 & 0.885 & \textbf{0.954} & 0.929 \\        
\hline
\end{tabular}}
\end{table}

The discussion of the results is divided into two parts: first, we analyze the learning performance on binary datasets, followed by an evaluation on multiclass datasets. Subsequently, we examine the properties of our algorithm in terms of communication and computation efficiency, scalability, impact of the chosen topology, and robustness to the presence of failures/malicious clients.
Fig.~\ref{fig:critdd_binary} and Fig.~\ref{fig:critdd_multiclass} compare DROCKS with all the competitors on binary and multiclass datasets, respectively. Tab.~\ref{tab:excerpt} provides an excerpt of the numerical results (f1-scores) on a small subset of datasets of the UCR archive. In particular, it shows a comparison between DROCKS and the competitors in terms of \textbf{F1-scores}. For the methods working with Rocket kernels, we present the results for the optimal number of kernels. Results (mean) are obtained with five averaged runs.

\textbf{Learning performance.}
On binary problems, FROCKS 
achieves the best performance, aligning with expectations, as the method was specifically designed for binary tasks. DROCKS consistently outperforms RawData, RocketFL, and the DL models (InceptionTime and ResNet). 

Overall, the critical diagram underscores the robustness of DROCKS and FROCKS in binary classification tasks. 
The RocketFL approach performs even worse than using raw data. We hypothesize that this occurs because each local client may rely on a distinct set of feature transformations depending on its specific data distribution. Without a mechanism to align these transformations across clients, the model fails to converge toward a consistent and effective feature representation, thus leading to performance degradation.
Moreover, the performance of DL methods suggests that kernel-based approaches are competitive in this domain.


As shown in Fig.~\ref{fig:critdd_multiclass}, DROCKS achieves the best performance in multiclass classification, significantly outperforming the competing approaches. This demonstrates its effectiveness in handling multiclass classification tasks compared to server-based methods, thus highlighting one of its key advantages.
An interesting observation is the impact of the number of kernels on performance. Fig.~\ref{fig:critdd_drocks} shows the results taking into account both binary and multiclass datasets. For DROCKS, the performance generally improves as the number of kernels increases (except for $K=10~000$), indicating its ability to leverage richer feature representations. 
We hypothesize that when adopting a huge number of kernels, the model may struggle to converge to a local minimum before being passed to another node for further training. This difficulty may arise primarily due to the high dimensionality of the model, which inherently increases the complexity of the optimization landscape. Consequently, the model may require more epochs and additional gradient descent steps to explore the parameter space and achieve meaningful convergence adequately.
The same anomaly is identified in RocketFL, which benefits from increasing the number of kernels but suffers with $K=10~000$, thus suggesting overfitting. 
A similar trend can be observed for FROCKS and RocketFL (results are shown in the Appendices~\ref{secA1:ablation} and~\ref{secA2:ablation}), although some anomalies are present: RocketFL with $500$ and $5~000$ kernels slightly outperforms configurations with $1~000$ and $10~000$ kernels, respectively. These variations could hint at overfitting or inefficiencies in leveraging the additional kernels. 
Similarly to the binary case, the RawData approach outperforms RocketFL, where the same set of ROCKET kernels is shared among all the clients. We hypothesize that this occurs because each client's local data may require different feature transformations, preventing convergence to a common set of features when using a uniform kernel distribution. 

DROCKS achieves higher accuracy than the baselines, with low standard deviations (Tables~\ref{tab:abl:binary_drocks} and~\ref{tab:abl:multi_drocks}). This means that the model performs well on all the nodes, leading to reliable and consistent decisions. DROCKS benefits from the exchange of local best-performing kernels, thus adapting the model to the current task while preserving features extracted from a previous node in a way that is reminiscent of experience replay techniques. This suggests that the feature extraction process benefits from exchanging the most effective kernels and continuously adapting them to the local data until the reach of a consensus among all the parties of the federation. This method is proposed as a promising alternative to the classic parameter averaging.

\begin{figure}[ht]
    \centering
    \begin{minipage}[b]{0.48\textwidth}
        \centering
        \includegraphics[width=0.97\columnwidth]{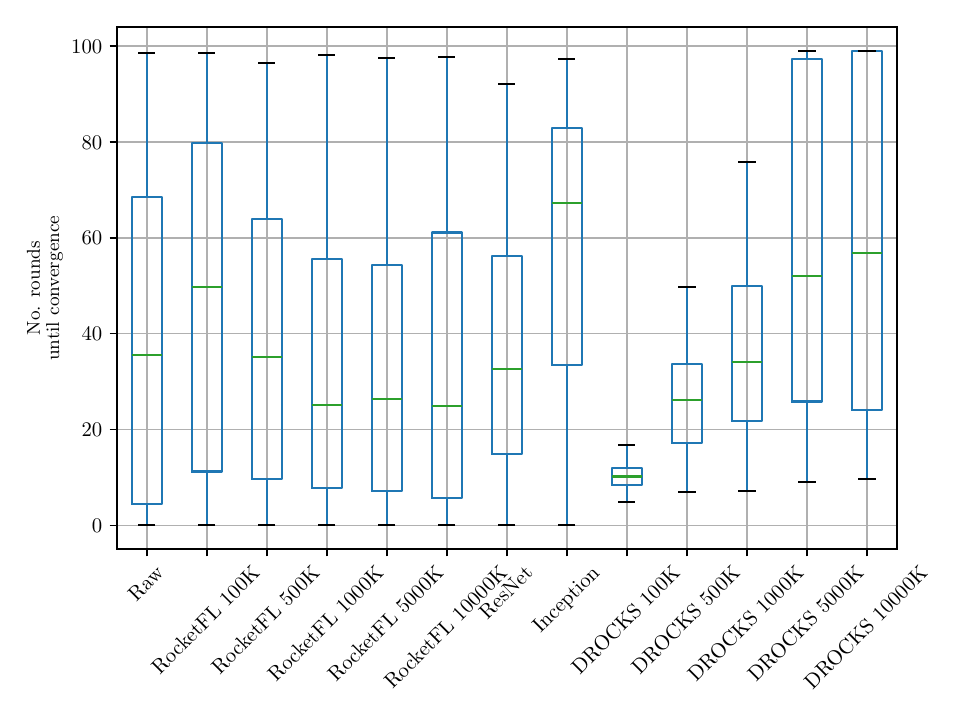} 
        \caption{Number of rounds until convergence shown over all datasets. The x-axis indicates the number of kernels used to initialize.}
        \label{fig:rounds_convergence}
    \end{minipage}
    \hfill
    \begin{minipage}[b]{0.48\textwidth}
        \centering
        \includegraphics[width=0.97\columnwidth]{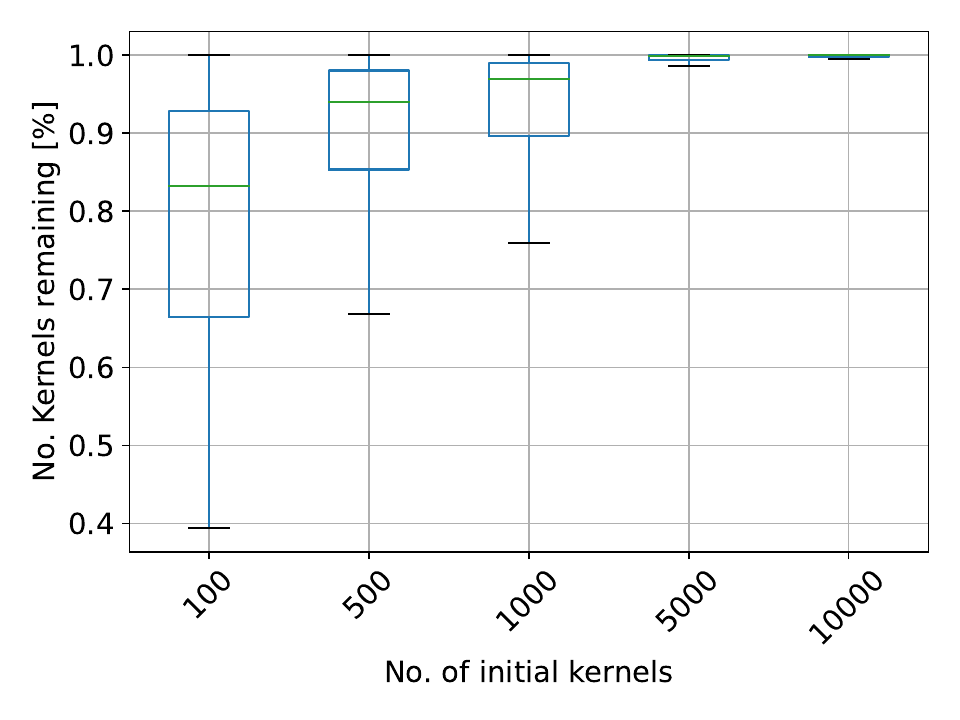} 
        \caption{The percentage of DROCKS' remaining kernels. The x-axis indicates the number of kernels used to initialize.}
        \label{fig:kernels_convergence}
    \end{minipage}
\end{figure}

\textbf{Analysis of communication and computational demand.}
Figure~\ref{fig:rounds_convergence} shows that DROCKS requires fewer rounds for convergence when dealing with a relatively small number of kernels. This is probably due to the federation's ability to reach an agreement on which features are the most effective transformations, whereas the RocketFL approach distributed all the features at once, thus slowing the consensus process. Additionally, thanks to its fast convergence properties, DROCKS demands less computation and communication resources, with training ending as soon as convergence is achieved. 
From Fig.~\ref{fig:kernels_convergence}, it can be seen that DROCKS will remove a portion of the initial features that are deemed unnecessary. This further reduces computation costs since the final model will have a smaller size than the initial model. It can be seen that the difficulty in reaching a consensus among which kernels are meaningful increases with the number of kernels used.

Eventually, as discussed in Sec.~\ref{sec:method}, DROCKS further reduces communication and computation overhead by requiring the training and transmission of only a single model per round, in contrast with common FL solutions requiring $N$ models. Additionally, compared to DL methods, DROCKS requires a fraction of the cost of transferring the model. Table~\ref{tab:abl:modelsizes} reports the statistics of the models in terms of sizes (megabytes) and number of parameters. 
DROCKS trains a logistic regression where the number of coefficients is relatively small compared to DL methods (thousands to billions of parameters). Indeed, a logistic regression model's number of parameters depends on the number of input features and output classes, with the addition of the bias terms (intercepts). 

\begin{table}[htbp]
\centering
\caption{Statistics of the models in terms of model sizes (megabytes) and number of parameters for a binary classification problem.}
\label{tab:abl:modelsizes}
\setlength{\tabcolsep}{1.1mm} 
\resizebox{\columnwidth}{!}{
\begin{tabular}{lccccc}
\hline
 & ResNet & InceptionTime & \multicolumn{3}{c}{Logistic Regression} \\
 & & & \multicolumn{3}{c}{Features} \\
 & & &  100 & 1~000 & 10~000 \\
 \hline
Model size (MB) & 14.74 & $4.3\mathrm{e}{-01}$ & $7.70\mathrm{e}{-04}$ & $7.63\mathrm{e}{-03}$ & $7.63\mathrm{e}{-02}$\\
Number of parameters & 3~853~834 & 110~794 & 101 & 1~001 & 10~001\\
\hline
\end{tabular}}
\end{table}

\begin{figure}[t]
\centering
    \includegraphics[width=0.99\columnwidth]{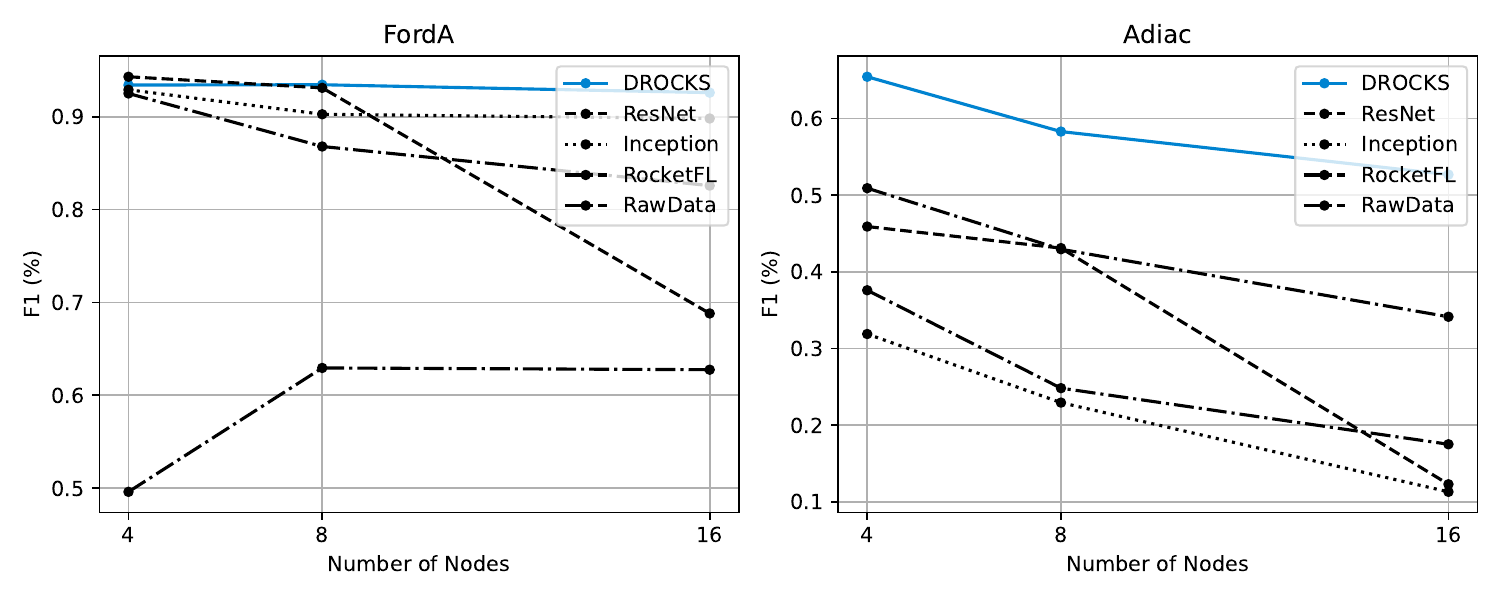} 
    \caption{Scalability performance w.r.t. number of nodes for the proposed approach and state-of-the-art methods.}
    \label{fig:scalability}
\end{figure}

\textbf{Scalability.}
We evaluate the DROCKS' capability to scale with the size of the federation. We tested this property using two random datasets from the UCR archive, e.g., FordA and Adiac, and by splitting them, with an i.i.d. setting, among an increasing number of federation participants. As a result, each node will hold a smaller portion of data. Fig.~\ref{fig:scalability} shows that DROCKS maintains high classification scores while competitors' performance drops more rapidly. This highlights its superior ability to handle increasing numbers of clients, preserving model accuracy where other methods degrade.

\begin{figure}[ht]
    \centering
    \begin{minipage}[b]{0.47\textwidth}
        \centering
        \includegraphics[width=0.9\columnwidth]{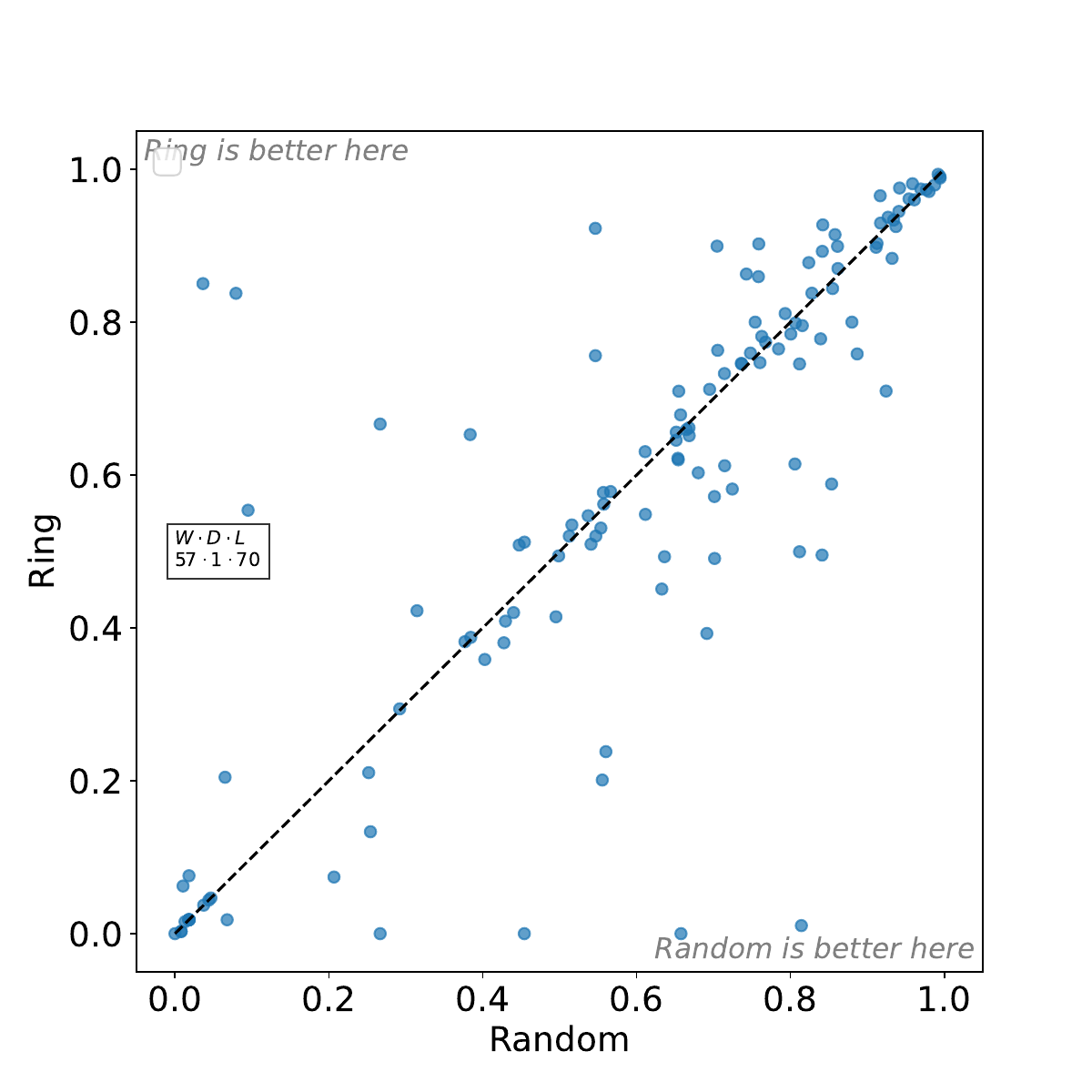} 
        \caption{Pairwise f1-score with cyclic model transfer (ring) versus considering a random node as subsequent.}
        \label{fig:random_topology_1000}
                \end{minipage}
    \hfill
    \begin{minipage}[b]{0.47\textwidth}
        \centering
        \begin{tabular}{lccc}
        \hline
        Dataset & \multicolumn{3}{c}{F1-Score} \\
         & \multicolumn{3}{c}{Remaining clients} \\
         & $100\%$ & $75\%$ & $50\%$ \\
         \hline
        ACSF1 & 0.593 & 0.621 & 0.619\\
        Adiac & 0.564 & 0.522 & 0.472\\
        ArrowHead & 0.635 & 0.636 & 0.590\\
        Beef & 0.498 & 0.462 & 0.403\\
        BeetleFly & 0.739 & 0.351 & 0.273\\
        \hline
        \end{tabular}
        \caption{Comparison of DROCKS with all clients vs. a reduced federation (due to malicious, compromised, or disconnected clients) in terms of F1-score on UCR datasets. 
        }
        \label{tab:abl:malicious_clients}
    \end{minipage}
\end{figure}

\textbf{Impact of the topology and of problematic clients.}
From the system's viewpoint, for each round of the training, DROCKS substitutes the master-worker schema with a pipeline schema for each iteration. Instead of delegating the reduction of the local models to a dedicated node (the master/server), each federation node works in a pipeline with other nodes along a path traversing all the nodes. Both schemas are equivalent in terms of functional semantics~\cite{danelutto1999stream}, whereas they exhibit different extra-functional characteristics. Although the pipeline is not a resilient schema since the failure of each of the nodes leads to the failure of the whole process, 
it does not suffer from knowledge asymmetry as no nodes have more information than each other.
We introduced fault tolerance to the system with a StreamFlow~\cite{colonnelli2021streamflow} implementation of DROCKS. StreamFlow introduces resilience properties, since the failure of each of the nodes does not lead to the failure of the whole process~\cite{mulone2024fault}. The continuity of the pipeline is maintained by managing crashes dynamically and by ensuring that the federation remains robust despite potential failures.
Moreover, the pipeline schema can be extended to dynamically bypass the failed/untrusted node (shortcutting the path), follow multiple paths along a direct acyclic graph, or consider a random node as a successor. Considering failures or even multiple paths will make the final model dependent on node and connectivity status (as shown in~\cite{pennisi2024feder}), shifting the target from a single global model to multiple (possibly similar) possible models, as it happens in the cross-device scenario. 
We show DROCKS' ability to handle different communication strategies and to deal with failed/malicious clients. We tested DROCKS considering a topology with a random node as a subsequent node (Fig.~\ref{fig:random_topology_1000}) and when removing one or more untrusted clients from the federation (Tab.~\ref{tab:abl:malicious_clients}). 
For these experiments, we considered that one or two random clients were excluded from the federation after five rounds of training. Final results are obtained by testing the final model on all data, including that of the excluded clients. For the topology experiments, we fixed the number of kernels to 1000 (additional results are available in Appendix~\ref{secA5:topology}), while for the experiments on dropping malicious clients, we fixed the number of kernels to 100 for simplicity.
Results (mean) are obtained with five averaged runs.

Fig.~\ref{fig:random_topology_1000} shows the pairwise f1-score of a ring communication schema versus considering a random client as a subsequent node for all the 128 datasets. Overall, they achieve similar results, with the ring topology being more precise than the random schema on 57 datasets and less precise on 70 datasets. However, for most of the datasets, the differences are pretty small. This suggests that, in terms of performance, the two topologies are roughly equivalent.

Table~\ref{tab:abl:malicious_clients} shows the f1-scores when removing one or more clients from the federation. In this set of experiments, we assumed that the untrusted clients would be removed after five training rounds. However, the final results are obtained by testing the resulting model on all clients' data, including the failed nodes. Intuitively, the performance decreases with the number of clients as the model is trained on less data.
When only one client is excluded from the federation, the performance is slightly decreased with respect to the 4-clients scenario. A considerable loss in performance happens when two clients are removed. However, some exceptions can happen, as in the case of the ACSF1 dataset, in which performance is even better with fewer clients.

\subsection{Conclusion}
\label{ssec:conclusion}

This work presents DROCKS, a decentralized FL approach for TSC based on ROCKET kernels. In our proposed method, each federation client sequentially trains the global model and contributes to the kernel selection step. In particular, each node in the sequence receives the trained model and the best-performing kernels (selected according to the largest squared weight associated) from the previous client. It then trains the model locally using a combination of the received kernels and a new set of ROCKET kernels. Results, spanning over 128 datasets of the UCR archive, show that our method outperforms state-of-the-art FedAvg-based approaches in multiclass classification. For binary classification tasks, DROCKS is slightly outperformed only by FROCKS, a method specifically designed for such scenarios.
Additionally, DROCKS significantly reduces computation and communication overhead: it converges in fewer rounds than the baselines and requires only half the communication per round compared to typical server-based methods. 
For future work, we aim to mitigate the performance drop observed when a large number of kernels is used.
Potential solutions may involve more gradient descent steps before model exchange, or exploiting a second-order solver, such as the Quasi-Newton method L-BFGS, to accelerate convergence.
Lastly, we plan to evaluate our approach in non-i.i.d. settings.

\begin{credits}
\subsubsection{\ackname} This research has been partly funded by the Federal Ministry of Education and Research of Germany and the state of North Rhine-Westphalia as part of the Lamarr Institute for Machine Learning and Artificial Intelligence, and partly supported by the Spoke "FutureHPC \& BigData" of the ICSC - Centro Nazionale di Ricerca in "High Performance Computing, Big Data and Quantum Computing", funded by European Union - NextGenerationEU, and by the Horizon2020 RIA EPI project (G.A. 826647).

\subsubsection{\discintname}
The authors have no competing interests to declare that are relevant to the content of this article. 
\end{credits}
%
%
%
\clearpage
\bibliographystyle{splncs04}
\bibliography{bibliography}
%
\clearpage
\section{Results for binary Datasets}
\label{secA1}

\begin{table*}[!ht]
\centering
\caption{Comparison between DROCKS and the competitors in terms of \textbf{F1-scores} on binary datasets. For the methods working with Rocket kernels, we present the results for the optimal number of kernels (i.e., RocketFL with 5~000 kernels, FROCKS with 10~000 kernels, and DROCKS with 5~000 kernels). Results (mean) are obtained with five averaged runs.}
\label{tab:binary_f1s}
\setlength{\tabcolsep}{1.1mm} 
\resizebox{.85\textwidth}{!}{
\begin{tabular}{lcccccc}
\hline
Dataset & RawData & ResNet & InceptioTime & RocketFL & FROCKS & DROCKS \\
 \hline   
        BeetleFly & 0.833 & 0.820 & 0.610 & 0.867 & 0.735 & 0.814 \\
        BirdChicken & 0.558 & 0.605 & 0.880 & 0.607 & 0.680 & 0.616 \\
        Chinatown & 0.725 & 0.726 & 0.724 & 0.724 & 0.982 & 0.923 \\
        Coffee & 0.250 & 0.178 & 0.105 & 0.118 & 0.940 & 0.889 \\
        Computers & 0.452 & 0.453 & 0.452 & 0.452 & 0.724 & 0.752 \\
        DistalPhalanxOutlineCorrect & 0.745 & 0.824 & 0.791 & 0.777 & 0.810 & 0.809 \\
        DodgerLoopGame & 0.458 & 0.458 & 0.458 & 0.458 & 0.673 & 0.635 \\
        DodgerLoopWeekend & 0.250 & 0.242 & 0.228 & 0.221 & 0.933 & 0.000 \\
        Earthquakes & 0.385 & 0.385 & 0.418 & 0.475 & 0.033 & 0.000 \\
        ECG200 & 0.797 & 0.875 & 0.834 & 0.867 & 0.866 & 0.854 \\
        ECGFiveDays & 0.823 & 0.644 & 0.649 & 0.689 & 0.819 & 0.689 \\
        FordA & 0.632 & 0.924 & 0.904 & 0.901 & 0.471 & 0.931 \\
        FordB & 0.654 & 0.790 & 0.756 & 0.756 & 0.111 & 0.794 \\
        FreezerRegularTrain & 0.502 & 0.502 & 0.502 & 0.502 & 0.834 & 0.964 \\
        FreezerSmallTrain & 0.500 & 0.500 & 0.500 & 0.500 & 0.805 & 0.811 \\
        GunPoint & 0.693 & 0.855 & 0.838 & 0.679 & 0.963 & 0.881 \\
        GunPointAgeSpan & 0.444 & 0.444 & 0.444 & 0.444 & 0.928 & 0.934 \\
        GunPointMaleVersusFemale & 0.436 & 0.444 & 0.442 & 0.441 & 0.987 & 0.991 \\
        GunPointOldVersusYoung & 0.556 & 0.556 & 0.556 & 0.517 & 0.966 & 0.970 \\
        Ham & 0.553 & 0.534 & 0.528 & 0.525 & 0.736 & 0.689 \\
        HandOutlines & 0.909 & 0.865 & 0.810 & 0.905 & 0.832 & 0.931 \\
        Herring & 0.630 & 0.559 & 0.551 & 0.554 & 0.516 & 0.375 \\
        HouseTwenty & 0.305 & 0.321 & 0.320 & 0.305 & 0.812 & 0.896 \\
        ItalyPowerDemand & 0.908 & 0.929 & 0.883 & 0.885 & 0.954 & 0.929 \\
        Lightning2 & 0.754 & 0.784 & 0.779 & 0.751 & 0.780 & 0.708 \\
        MiddlePhalanxOutlineCorrect & 0.708 & 0.820 & 0.771 & 0.774 & 0.813 & 0.833 \\
        MoteStrain & 0.722 & 0.634 & 0.691 & 0.616 & 0.925 & 0.821 \\
        PhalangesOutlinesCorrect & 0.759 & 0.837 & 0.789 & 0.751 & 0.787 & 0.843 \\
        PowerCons & 0.879 & 0.831 & 0.790 & 0.613 & 0.929 & 0.936 \\
        ProximalPhalanxOutlineCorrect & 0.842 & 0.898 & 0.818 & 0.793 & 0.891 & 0.820 \\
        SemgHandGenderCh2 & 0.328 & 0.328 & 0.328 & 0.328 & 0.334 & 0.588 \\
        ShapeletSim & 0.503 & 0.503 & 0.507 & 0.503 & 0.626 & 0.133 \\
        SonyAIBORobotSurface1 & 0.657 & 0.770 & 0.666 & 0.562 & 0.715 & 0.705 \\
        SonyAIBORobotSurface2 & 0.788 & 0.801 & 0.760 & 0.733 & 0.839 & 0.844 \\
        Strawberry & 0.825 & 0.859 & 0.796 & 0.830 & 0.949 & 0.933 \\
        ToeSegmentation1 & 0.486 & 0.487 & 0.486 & 0.491 & 0.823 & 0.664 \\
        ToeSegmentation2 & 0.203 & 0.208 & 0.203 & 0.203 & 0.660 & 0.586 \\
        TwoLeadECG & 0.680 & 0.632 & 0.652 & 0.632 & 0.878 & 0.772 \\
        Wafer & 0.970 & 0.996 & 0.995 & 0.973 & 0.993 & 0.991 \\
        Wine & 0.741 & 0.383 & 0.375 & 0.447 & 0.688 & 0.400 \\
        WormsTwoClass & 0.491 & 0.492 & 0.495 & 0.491 & 0.770 & 0.763 \\
        Yoga & 0.679 & 0.715 & 0.706 & 0.706 & 0.797 & 0.750 \\
\hline
\end{tabular}}
\end{table*}

\begin{table*}[!ht]
\centering
\caption{Comparison between DROCKS and the competitors in terms of \textbf{accuracy} on binary datasets. For the methods working with Rocket kernels, we present the results for the optimal number of kernels (i.e., RocketFL with 5~000 kernels, FROCKS with 10~000 kernels, and DROCKS with 5~000 kernels). Results (mean) are obtained with five averaged runs.}
\label{tab:binary_accuracy}
\setlength{\tabcolsep}{1.1mm} 
\resizebox{.85\textwidth}{!}{
\begin{tabular}{lcccccc}
\hline
Dataset & RawData & ResNet & InceptioTime & RocketFL & FROCKS & DROCKS \\
 \hline
BeetleFly & 0.888 & 0.888 & 0.663 & 0.913 & 0.800 & 0.780 \\
        BirdChicken & 0.713 & 0.788 & 0.850 & 0.688 & 0.700 & 0.610 \\
        Chinatown & 0.968 & 0.922 & 0.813 & 0.884 & 0.974 & 0.893 \\
        Coffee & 1.000 & 0.638 & 0.363 & 0.450 & 0.943 & 0.871 \\
        Computers & 0.538 & 0.677 & 0.768 & 0.697 & 0.732 & 0.743 \\
        DistalPhalanxOutlineCorrect & 0.687 & 0.791 & 0.758 & 0.763 & 0.763 & 0.770 \\
        DodgerLoopGame & 0.788 & 0.592 & 0.545 & 0.713 & 0.712 & 0.552 \\
        DodgerLoopWeekend & 0.976 & 0.901 & 0.717 & 0.903 & 0.965 & 0.739 \\
        Earthquakes & 0.622 & 0.762 & 0.769 & 0.615 & 0.753 & 0.748 \\
        ECG200 & 0.788 & 0.844 & 0.771 & 0.848 & 0.826 & 0.806 \\
        ECGFiveDays & 0.866 & 0.583 & 0.626 & 0.746 & 0.846 & 0.788 \\
        FordA & 0.496 & 0.943 & 0.929 & 0.925 & 0.683 & 0.934 \\
        FordB & 0.517 & 0.821 & 0.787 & 0.785 & 0.525 & 0.776 \\
        FreezerRegularTrain & 0.927 & 0.917 & 0.913 & 0.940 & 0.860 & 0.965 \\
        FreezerSmallTrain & 0.779 & 0.733 & 0.709 & 0.758 & 0.814 & 0.827 \\
        GunPoint & 0.773 & 0.888 & 0.847 & 0.736 & 0.964 & 0.871 \\
        GunPointAgeSpan & 0.901 & 0.944 & 0.828 & 0.899 & 0.933 & 0.937 \\
        GunPointMaleVersusFemale & 0.914 & 0.989 & 0.981 & 0.980 & 0.988 & 0.992 \\
        GunPointOldVersusYoung & 1.000 & 1.000 & 1.000 & 0.950 & 0.964 & 0.968 \\
        Ham & 0.737 & 0.632 & 0.648 & 0.593 & 0.712 & 0.682 \\
        HandOutlines & 0.886 & 0.825 & 0.732 & 0.893 & 0.817 & 0.912 \\
        Herring & 0.681 & 0.600 & 0.616 & 0.588 & 0.650 & 0.556 \\
        HouseTwenty & 0.757 & 0.927 & 0.908 & 0.883 & 0.835 & 0.916 \\
        ItalyPowerDemand & 0.937 & 0.951 & 0.906 & 0.904 & 0.955 & 0.933 \\
        Lightning2 & 0.693 & 0.686 & 0.674 & 0.674 & 0.731 & 0.669 \\
        MiddlePhalanxOutlineCorrect & 0.626 & 0.799 & 0.738 & 0.758 & 0.760 & 0.797 \\
        MoteStrain & 0.848 & 0.780 & 0.807 & 0.713 & 0.929 & 0.843 \\
        PhalangesOutlinesCorrect & 0.678 & 0.819 & 0.757 & 0.756 & 0.733 & 0.792 \\
        PowerCons & 0.969 & 0.929 & 0.906 & 0.824 & 0.933 & 0.937 \\
        ProximalPhalanxOutlineCorrect & 0.848 & 0.904 & 0.798 & 0.792 & 0.846 & 0.773 \\
        SemgHandGenderCh2 & 0.866 & 0.668 & 0.751 & 0.801 & 0.721 & 0.660 \\
        ShapeletSim & 0.493 & 0.586 & 0.615 & 0.774 & 0.698 & 0.500 \\
        SonyAIBORobotSurface1 & 0.757 & 0.837 & 0.640 & 0.713 & 0.658 & 0.643 \\
        SonyAIBORobotSurface2 & 0.808 & 0.798 & 0.701 & 0.713 & 0.813 & 0.793 \\
        Strawberry & 0.917 & 0.964 & 0.879 & 0.935 & 0.935 & 0.914 \\
        ToeSegmentation1 & 0.567 & 0.654 & 0.597 & 0.827 & 0.840 & 0.711 \\
        ToeSegmentation2 & 0.557 & 0.836 & 0.839 & 0.837 & 0.874 & 0.808 \\
        TwoLeadECG & 0.768 & 0.652 & 0.616 & 0.664 & 0.888 & 0.798 \\
        Wafer & 0.950 & 0.993 & 0.991 & 0.956 & 0.988 & 0.983 \\
        Wine & 0.857 & 0.851 & 0.864 & 0.673 & 0.641 & 0.500 \\
        WormsTwoClass & 0.585 & 0.841 & 0.803 & 0.705 & 0.704 & 0.735 \\
        Yoga & 0.626 & 0.701 & 0.705 & 0.714 & 0.750 & 0.733 \\

\hline
\end{tabular}}
\end{table*}

\clearpage
\subsection{Ablation study on ROCKET kernels}
\label{secA1:ablation}

\begin{table*}[!ht]
\centering
\caption{Performance in terms of accuracy and F1-scores of the \textbf{RocketFL} baseline w.r.t. the number of kernels (100, 500, 1\,000, 5\,000, and 10\,000). Results (mean) are obtained with five averaged runs.}
\label{tab:abl:binary_rocket}
\setlength{\tabcolsep}{1.1mm} 
\resizebox{.87\textwidth}{!}{
\begin{tabular}{lcccccccccc}
\hline
\multicolumn{1}{l}{Dataset} & \multicolumn{5}{c}{Accuracy} & \multicolumn{5}{c}{F1-Score} \\
 & 100 & 500 & 1000 & 5000 & 10000 & 100 & 500 & 1000 & 5000 & 10000 \\
 \hline
        BeetleFly & 0.800 & 0.875 & 0.875 & 0.913 & 0.875 & 0.680 & 0.800 & 0.800 & 0.867 & 0.783 \\ 
        BirdChicken & 0.650 & 0.700 & 0.713 & 0.688 & 0.713 & 0.543 & 0.615 & 0.600 & 0.607 & 0.630 \\ 
        Chinatown & 0.869 & 0.866 & 0.871 & 0.884 & 0.878 & 0.724 & 0.724 & 0.724 & 0.724 & 0.724 \\ 
        Coffee & 0.475 & 0.438 & 0.438 & 0.450 & 0.488 & 0.118 & 0.123 & 0.113 & 0.118 & 0.118 \\ 
        Computers & 0.663 & 0.684 & 0.697 & 0.697 & 0.702 & 0.452 & 0.452 & 0.452 & 0.452 & 0.452 \\ 
        DistalPhalanxOutlineCorrect & 0.716 & 0.753 & 0.750 & 0.763 & 0.773 & 0.747 & 0.770 & 0.771 & 0.777 & 0.789 \\
\hline
\end{tabular}
}
\end{table*}

\begin{table*}[htbp]
\centering
\caption{Performance in terms of accuracy and F1-scores of \textbf{FROCKS} w.r.t. the number of kernels (100, 500, 1\,000, 5\,000, and 10\,000). Results (mean) are obtained with five averaged runs.}
\label{tab:abl:binary_frocks}
\setlength{\tabcolsep}{1.1mm} 
\resizebox{.87\textwidth}{!}{
\begin{tabular}{lcccccccccc}
\hline
\multicolumn{1}{l}{Dataset} & \multicolumn{5}{c}{Accuracy} & \multicolumn{5}{c}{F1-Score} \\
 & 100 & 500 & 1000 & 5000 & 10000 & 100 & 500 & 1000 & 5000 & 10000 \\
 \hline
BeetleFly & 0,750 & 0,800 & 0,730 & 0,850 & 0,800 & 0,627 & 0,698 & 0,637 & 0,791 & 0,735 \\
        BirdChicken & 0,660 & 0,640 & 0,690 & 0,700 & 0,700 & 0,514 & 0,709 & 0,676 & 0,782 & 0,680 \\
        Chinatown & 0,964 & 0,964 & 0,971 & 0,975 & 0,974 & 0,975 & 0,272 & 0,980 & 0,605 & 0,982 \\
        Coffee & 0,850 & 0,871 & 0,886 & 0,921 & 0,943 & 0,858 & 0,987 & 0,879 & 0,991 & 0,940 \\
        Computers & 0,708 & 0,705 & 0,708 & 0,723 & 0,732 & 0,692 & 0,677 & 0,689 & 0,835 & 0,724 \\
        DistalPhalanxOutlineCorrect & 0,672 & 0,745 & 0,738 & 0,773 & 0,763 & 0,763 & 0,564 & 0,803 & 0,653 & 0,810 \\
\hline
\end{tabular}
}
\end{table*}

\begin{table*}[!ht]
\centering
\caption{Performance in terms of accuracy and F1-scores of \textbf{DROCKS} w.r.t. the number of kernels (100, 500, 1\,000, 5\,000, and 10\,000). Results (mean) are obtained with five averaged runs.}
\label{tab:abl:binary_drocks}
\setlength{\tabcolsep}{1.1mm} 
\resizebox{.99\textwidth}{!}{
\begin{tabular}{lcccccccccc}
\hline
\multicolumn{1}{l}{Dataset} & \multicolumn{5}{c}{Accuracy} & \multicolumn{5}{c}{F1-Score} \\
 & 100 & 500 & 1000 & 5000 & 10000 & 100 & 500 & 1000 & 5000 & 10000 \\
 \hline
BeetleFly & $0.750 \pm 0.158$ & $0.800 \pm 0.106$ & $0.780 \pm 0.076$ & $0.780 \pm 0.076$ & $0.780 \pm 0.084$ & $0.739 \pm 0.166$ & $0.834 \pm 0.080$ & $0.812 \pm 0.047$ & $0.814 \pm 0.051$ & $0.814 \pm 0.059$ \\
BirdChicken & $0.620 \pm 0.135$ & $0.630 \pm 0.104$ & $0.650 \pm 0.079$ & $0.610 \pm 0.096$ & $0.620 \pm 0.104$ & $0.540 \pm 0.189$ & $0.619 \pm 0.121$ & $0.658 \pm 0.078$ & $0.616 \pm 0.105$ & $0.631 \pm 0.106$ \\
        Chinatown & $0.889 \pm 0.087$ & $0.897 \pm 0.084$ & $0.894 \pm 0.086$ & $0.893 \pm 0.087$ & $0.893 \pm 0.087$ & $0.927 \pm 0.056$ & $0.927 \pm 0.061$ & $0.924 \pm 0.063$ & $0.923 \pm 0.065$ & $0.923 \pm 0.065$ \\
        Coffee & $0.850 \pm 0.125$ & $0.936 \pm 0.069$ & $0.964 \pm 0.036$ & $0.871 \pm 0.197$ & $0.950 \pm 0.041$ & $0.781 \pm 0.237$ & $0.922 \pm 0.093$ & $0.961 \pm 0.039$ & $0.889 \pm 0.145$ & $0.948 \pm 0.040$ \\
        Computers & $0.663 \pm 0.074$ & $0.719 \pm 0.039$ & $0.722 \pm 0.043$ & $0.743 \pm 0.032$ & $0.738 \pm 0.028$ & $0.663 \pm 0.074$ & $0.740 \pm 0.027$ & $0.737 \pm 0.030$ & $0.752 \pm 0.034$ & $0.749 \pm 0.029$ \\
        DistalPhalanxOutlineCorrect & $0.713 \pm 0.017$ & $0.748 \pm 0.012$ & $0.747 \pm 0.012$ & $0.770 \pm 0.007$ & $0.742 \pm 0.025$ & $0.766 \pm 0.009$ & $0.798 \pm 0.010$ & $0.801 \pm 0.015$ & $0.809 \pm 0.012$ & $0.783 \pm 0.034$ \\
\hline
\end{tabular}
}
\end{table*}

\clearpage
\section{Results for multiclass Datasets}
\label{secA2}

\setlength{\tabcolsep}{1.1mm}

\onecolumn
\begin{longtable}{lccccc}
\caption{Comparison between DROCKS and the competitors in terms of \textbf{F1-scores} on multiclass datasets. For the methods working with Rocket kernels, we present the results for the optimal number of kernels (i.e., both RocketFL and DROCKS with 5~000). Results (mean) are obtained with five averaged runs.}
\label{tab:multiclass_f1s} \\
\hline
Dataset & RawData & ResNet & InceptioTime & RocketFL & DROCKS \\
\hline
\endfirsthead
\caption*{(Continued)} \\

\hline
Dataset & RawData & ResNet & InceptioTime & RocketFL & DROCKS \\
\hline
\endhead

\hline
\multicolumn{6}{r}{(Continued on next page)} \\
\endfoot

\hline
\endlastfoot

        ACSF1 & 0.498 & 0.527 & 0.597 & 0.716 & 0.640 \\
        Adiac & 0.509 & 0.459 & 0.319 & 0.376 & 0.652 \\
        AllGestureWiimoteX & 0.108 & 0.215 & 0.206 & 0.296 & 0.018 \\
        AllGestureWiimoteY & 0.147 & 0.165 & 0.179 & 0.229 & 0.018 \\
        AllGestureWiimoteZ & 0.104 & 0.209 & 0.181 & 0.257 & 0.018 \\
        ArrowHead & 0.509 & 0.427 & 0.439 & 0.427 & 0.69 \\
        Beef & 0.608 & 0.532 & 0.405 & 0.693 & 0.523 \\
        BME & 0.745 & 0.447 & 0.398 & 0.528 & 0.872 \\
\end{longtable}
\clearpage
\twocolumn

\onecolumn
\begin{longtable}{lccccc}
\caption{Comparison between DROCKS and the competitors in terms of \textbf{accuracy} on multiclass datasets. For the methods working with Rocket kernels, we present the results for the optimal number of kernels (i.e., both RocketFL and DROCKS with 5~000 kernels). Results (mean) are obtained with five averaged runs.}
\label{tab:multiclass_accuracies} \\
\hline
Dataset & RawData & ResNet & InceptioTime & RocketFL & DROCKS \\
\hline
\endfirsthead
\caption*{(Continued)} \\

\hline
Dataset & RawData & ResNet & InceptioTime & RocketFL & DROCKS \\
\hline
\endhead

\hline
\multicolumn{6}{r}{(Continued on next page)} \\
\endfoot

\hline
\endlastfoot

ACSF1 & 0.623 & 0.663 & 0.704 & 0.827 & 0,662 \\
        Adiac & 0.647 & 0.595 & 0.452 & 0.511 & 0,679 \\
        AllGestureWiimoteX & 0.257 & 0.407 & 0.356 & 0.543 & 0,1 \\
        AllGestureWiimoteY & 0.297 & 0.343 & 0.362 & 0.500 & 0,1 \\
        AllGestureWiimoteZ & 0.234 & 0.360 & 0.342 & 0.482 & 0,1 \\
        ArrowHead & 0.737 & 0.547 & 0.504 & 0.541 & 0,696 \\
        Beef & 0.663 & 0.600 & 0.485 & 0.750 & 0,533 \\
        BME & 0.831 & 0.570 & 0.482 & 0.676 & 0,879 \\
\end{longtable}
\clearpage
\twocolumn

\newpage
\subsection{Ablation study on ROCKET kernels}
\label{secA2:ablation}

\onecolumn
\begin{longtable}{lcccccccccc}
\caption{Performance in terms of accuracy and F1-scores of the \textbf{RocketFL} baseline on multiclass datasets w.r.t. the number of kernels (100, 500, 1\,000, 5\,000, and 10\,000). Results (mean) are obtained with five averaged runs.} \\
\hline
\multicolumn{1}{l}{Dataset} & \multicolumn{5}{c}{Accuracy} & \multicolumn{5}{c}{F1-Score} \\
 & 100 & 500 & 1000 & 5000 & 10000 & 100 & 500 & 1000 & 5000 & 10000 \\
\hline
\endfirsthead
\caption*{(Continued)} \\
\hline
\multicolumn{1}{l}{Dataset} & \multicolumn{5}{c}{Accuracy} & \multicolumn{5}{c}{F1-Score} \\
 & 100 & 500 & 1000 & 5000 & 10000 & 100 & 500 & 1000 & 5000 & 10000 \\
\hline
\endhead
\hline
\multicolumn{11}{r}{(Continued on next page)} \\
\endfoot
\hline
\endlastfoot
ACSF1 & 0.746 & 0.796 & 0.829 & 0.827 & 0.802 & 0.610 & 0.655 & 0.715 & 0.716 & 0.696 \\
        Adiac & 0.441 & 0.510 & 0.516 & 0.511 & 0.479 & 0.312 & 0.378 & 0.386 & 0.376 & 0.344 \\
        AllGestureWiimoteX & 0.449 & 0.519 & 0.515 & 0.543 & 0.535 & 0.219 & 0.261 & 0.266 & 0.296 & 0.287 \\
        AllGestureWiimoteY & 0.412 & 0.476 & 0.483 & 0.500 & 0.488 & 0.183 & 0.212 & 0.219 & 0.229 & 0.224 \\
        AllGestureWiimoteZ & 0.388 & 0.442 & 0.470 & 0.482 & 0.483 & 0.193 & 0.232 & 0.246 & 0.257 & 0.255 \\
        ArrowHead & 0.520 & 0.531 & 0.527 & 0.541 & 0.536 & 0.427 & 0.427 & 0.427 & 0.427 & 0.427 \\
        Beef & 0.747 & 0.772 & 0.767 & 0.750 & 0.762 & 0.687 & 0.724 & 0.719 & 0.693 & 0.727 \\
        BeetleFly & 0.800 & 0.875 & 0.875 & 0.913 & 0.875 & 0.680 & 0.800 & 0.800 & 0.867 & 0.783 \\
\hline
\end{longtable}

\clearpage

\begin{landscape}
\scriptsize
\setlength{\tabcolsep}{0.4mm}
\begin{longtable}{lcccccccccc}
\caption{Performance in terms of accuracy and F1-scores of \textbf{DROCKS} on multiclass datasets w.r.t. the number of kernels (100, 500, 1\,000, 5\,000, and 10\,000). Results (mean) are obtained with five averaged runs. \label{tab:abl:multi_drocks}}\\

\hline
\multicolumn{1}{l}{Dataset} & \multicolumn{5}{c}{Accuracy} & \multicolumn{5}{c}{F1-Score} \\
 & 100 & 500 & 1000 & 5000 & 10000 & 100 & 500 & 1000 & 5000 & 10000 \\
 \hline
\endfirsthead
\caption*{(Continued)} \\
\hline
\multicolumn{1}{l}{Dataset} & \multicolumn{5}{c}{Accuracy} & \multicolumn{5}{c}{F1-Score} \\
 & 100 & 500 & 1000 & 5000 & 10000 & 100 & 500 & 1000 & 5000 & 10000 \\
\hline
\endhead
\hline
\multicolumn{11}{r}{(Continued on next page)} \\
\endfoot
\hline
\endlastfoot
ACSF1 & $0.618 \pm 0.051$ & $0.612 \pm 0.091$ & $0.654 \pm 0.092$ & $0.662 \pm 0.076$ & $0.668 \pm 0.076$ & $0.593 \pm 0.056$ & $0.592 \pm 0.104$ & $0.636 \pm 0.104$ & $0.640 \pm 0.092$ & $0.644 \pm 0.096$ \\
        Adiac & $0.593 \pm 0.018$ & $0.670 \pm 0.017$ & $0.678 \pm 0.015$ & $0.679 \pm 0.024$ & $0.629 \pm 0.058$ & $0.564 \pm 0.029$ & $0.644 \pm 0.021$ & $0.654 \pm 0.010$ & $0.652 \pm 0.022$ & $0.587 \pm 0.048$ \\
        AllGestureWiimoteX & $0.100 \pm 0.000$ & $0.100 \pm 0.000$ & $0.100 \pm 0.000$ & $0.100 \pm 0.000$ & $0.100 \pm 0.000$ & $0.018 \pm 0.000$ & $0.018 \pm 0.000$ & $0.018 \pm 0.000$ & $0.018 \pm 0.000$ & $0.018 \pm 0.000$ \\
        AllGestureWiimoteY & $0.100 \pm 0.000$ & $0.100 \pm 0.000$ & $0.100 \pm 0.000$ & $0.100 \pm 0.000$ & $0.100 \pm 0.000$ & $0.018 \pm 0.000$ & $0.018 \pm 0.000$ & $0.018 \pm 0.000$ & $0.018 \pm 0.000$ & $0.018 \pm 0.000$ \\
        AllGestureWiimoteZ & $0.100 \pm 0.000$ & $0.100 \pm 0.000$ & $0.100 \pm 0.000$ & $0.100 \pm 0.000$ & $0.100 \pm 0.000$ & $0.018 \pm 0.000$ & $0.018 \pm 0.000$ & $0.018 \pm 0.000$ & $0.018 \pm 0.000$ & $0.018 \pm 0.000$ \\
        ArrowHead & $0.643 \pm 0.056$ & $0.719 \pm 0.083$ & $0.706 \pm 0.088$ & $0.696 \pm 0.071$ & $0.693 \pm 0.078$ & $0.635 \pm 0.046$ & $0.714 \pm 0.076$ & $0.701 \pm 0.084$ & $0.690 \pm 0.067$ & $0.687 \pm 0.073$ \\
        Beef & $0.507 \pm 0.049$ & $0.527 \pm 0.092$ & $0.547 \pm 0.051$ & $0.533 \pm 0.041$ & $0.533 \pm 0.053$ & $0.498 \pm 0.059$ & $0.520 \pm 0.089$ & $0.546 \pm 0.050$ & $0.523 \pm 0.058$ & $0.524 \pm 0.067$ \\
        BME & $0.820 \pm 0.114$ & $0.849 \pm 0.098$ & $0.861 \pm 0.104$ & $0.879 \pm 0.099$ & $0.871 \pm 0.105$ & $0.806 \pm 0.129$ & $0.842 \pm 0.105$ & $0.853 \pm 0.112$ & $0.872 \pm 0.107$ & $0.862 \pm 0.114$ \\
\end{longtable}
\end{landscape}

\normalsize
\setlength{\tabcolsep}{1.1mm}

\clearpage
\section{FROCKS algorithm}
\label{secA4:frocks_multiclass}

\begin{algorithm}
\caption{FROCKS: Federated ROCKET FeatureS}
\label{alg:frocks}
\begin{algorithmic}[1]
\Require $C$: number of clients, $\{D_c = (X_c, y_c)\}_{c=1}^C$: local datasets, where $X_c$ are inputs and $y_c$ are labels for client $c$, $K$: number of ROCKET kernels for each client, $R$: number of training rounds
\Ensure Trained global model $w_{global}$ with selected best-performing kernels

\State Initialize a set of different $K$ kernels for each client based on consecutive seeding: $\{\mathcal{K}^{(0)} \}_{c=1}^C$, where $\mathcal{K}^{(0)} = \{k_{c,i}\}_{i=1}^{K}$, and where $k_{c,i}$ is initialized using a seed $s_{c,i} = (c-1)K+i-1$, with $c \in \{1,...,C\}$ and $i \in {1,...,K}$.
\State Initialize logistic regression models for all clients: $\{w_c^{(0)}\}_{c=1}^C$

\Function{$f_{\mathcal{K}_c}(X_c)$}{Kernel-based transformation of data}
    \For{each kernel $k \in \mathcal{K}_c$}
        \State Apply convolution: $y = k * x$ for each $x \in X_c$
        \State Compute PPV: $\text{PPV}(y) = \frac{1}{L} \sum_{i=1}^L \max(0, \text{sign}(y[i]))$, with $L$ being the length of the kernel output $y$.
    \EndFor
    \State Return feature matrix $Z_c$ with PPV values as features
\EndFunction

\For{$r = 0$ to $R$}
    \For{each client $c \in \{1, \dots, C\}$ in parallel}
        \State Transform local data $X_c$ using local kernels $\mathcal{K}^{(r)}$: $Z_c = f_{\mathcal{K}_c^{(r)}}(X_c)$

        \State Train the logistic regression model using the transformed data: $w_c^{(r)} = \text{argmin}_w \ell(w; Z_c, y_c)$
        
        \State Select the best-performing $p=\frac{K}{C}$ kernels based on weight magnitude:


        \[ p_c = \left[ w_{c,k}^{(r)}\right]_{k \in \mathcal{K}_c^{(r)}}, \ |w_{c,k}^{(r)}| \ is \ among \ top \ K \]
        
        \State Send the selected $p$ kernels and their weights $w_{c,k}^{(r)}$ to the server

    \EndFor

    \State Server aggregates kernels 
    
    \[ \mathcal{K}^{(r)} = \bigcup_{c=1}^C p_c \]
    
    and weights:
    \For{each unique kernel received from clients}
        \If{kernel is reported by multiple clients}
            \State Compute the average of the corresponding weights:
            \[ w_k^{(r)} = \begin{cases}
            \frac{1}{|\{c : k \in p_c\}|} \sum_{c : k \in p_c} w_{c,k}^{(r)} & \text{if } k \in \mathcal{K}^{(r)} \\
            0 & \text{otherwise}
    \end{cases} \]
        \EndIf
    \EndFor
    \State Server distributes the updated global kernels $\mathcal{K}^{(r)}$ and weights $\{w_k^{(r)}\}_{k \in \mathcal{K}^{(r)}}$ to all clients

\EndFor

\State Stop training when no significant change is detected in the kernels $\mathcal{K}^{(r)} = \mathcal{K}^{(r-1)}$ and weights $\|w^{(r)} - w^{(r-1)}\|_\infty < \epsilon$

\end{algorithmic}
\end{algorithm}

\newpage
\section{Additional results on the topology}
\label{secA5:topology}

\begin{figure}
\centering
    \includegraphics[width=0.9\columnwidth]{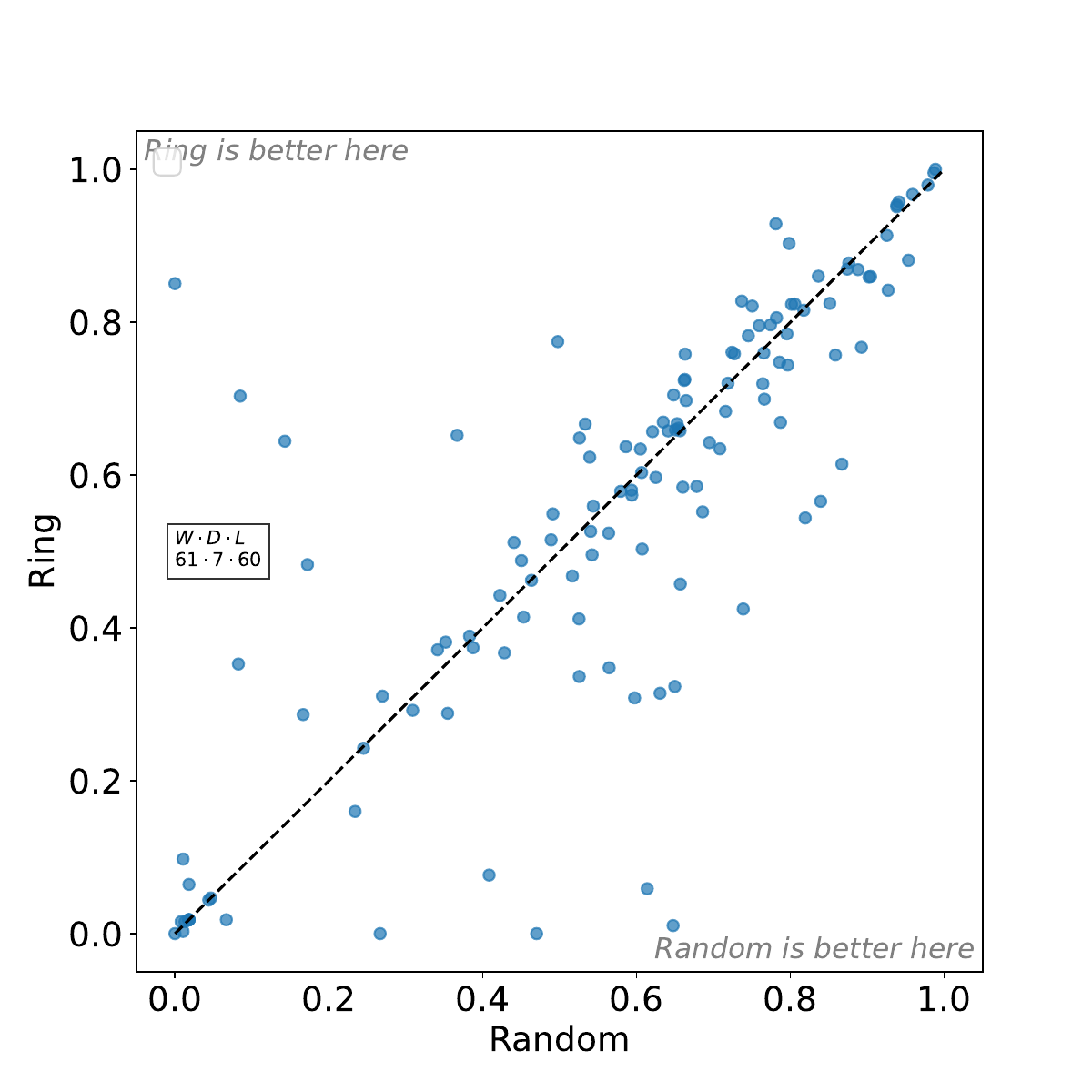} 
    \caption{Pairwise accuracy with cyclic model transfer (ring) versus considering a random node as subsequent with 100 kernels.}
    \label{fig:random_topology_100}
\end{figure}

\end{document}